\documentclass[10pt,journal,compsoc]{IEEEtran}
\usepackage{graphicx}
\usepackage{dsfont} 
\usepackage{amsmath}
\usepackage{amssymb}
\usepackage{booktabs}
\usepackage{bbm}
\usepackage{arydshln} 
\usepackage{epsfig}
\usepackage{makecell,rotating}
\usepackage{array}
\usepackage{colortbl}
\usepackage{xcolor}
\usepackage{multirow}
\usepackage{algorithm}  
\usepackage{subcaption}
\usepackage{algpseudocode}

\usepackage[pagebackref,breaklinks,colorlinks]{hyperref}
\usepackage{fancyvrb}
\usepackage{url}

\ifCLASSOPTIONcompsoc
  \usepackage[nocompress]{cite}
\else
  \usepackage{cite}
\fi

\ifCLASSINFOpdf
 
\else

\fi

\begin{document}

\title{Evolving Without Ending: \\Unifying Multimodal Incremental Learning for Continual Panoptic Perception}

\author{Bo~Yuan, 
       Danpei~Zhao*, 
       Wentao~Li,
       Tian~Li,
       Zhiguo~Jiang
\IEEEcompsocitemizethanks{\IEEEcompsocthanksitem  Bo Yuan, Danpei Zhao, Wentao Li,  Tian Li, Zhiguo Jiang are with the Department of Aerospace Intelligent Science and Technology, School of Astronautics, and the Key Laboratory of Spacecraft Design Optimization and Dynamic Simulation Technology, Ministry of Education, Beihang University, Beijing 102206, China, and also with the Tianmushan Laboratory, Hangzhou 311115, China.\protect\\
E-mail:  \{yuanbobuaa, zhaodanpei, canoe, lit, jiangzg\}@buaa.edu.cn \protect\\
* Corresponding author.
}
}

\markboth{Journal of \LaTeX\ Class Files,~Vol.~XX, No.~XX, XX~XX}%
{Shell \MakeLowercase{\textit{et al.}}: Bare Demo of IEEEtran.cls for Computer Society Journals}

\IEEEtitleabstractindextext{%
\begin{abstract}
Continual learning (CL) is a great endeavour in developing intelligent perception AI systems. However, the pioneer research has predominantly focus on single-task CL, which restricts the potential in multi-task and multimodal scenarios.  Beyond the well-known issue of catastrophic forgetting,  the multi-task CL also brings semantic obfuscation across multimodal alignment, leading to severe model degradation during incremental training steps. In this paper, we extend CL to continual panoptic perception (CPP), integrating multimodal and multi-task CL to enhance comprehensive image perception through pixel-level, instance-level, and image-level joint interpretation. We formalize the CL task in multimodal scenarios and propose an end-to-end continual panoptic perception model. Concretely, CPP model features a collaborative cross-modal encoder (CCE) for multimodal embedding. We also propose a malleable knowledge inheritance module via contrastive feature distillation and instance distillation, addressing catastrophic forgetting from task-interactive boosting manner. Furthermore, we propose a cross-modal consistency constraint and develop CPP+, ensuring multimodal semantic alignment for model updating under multi-task incremental scenarios. Additionally, our proposed model incorporates an asymmetric pseudo-labeling manner, enabling model evolving without exemplar replay. Extensive experiments on multimodal datasets and diverse CL tasks demonstrate the superiority of the proposed model, particularly in fine-grained CL tasks. 
\end{abstract}

}

\maketitle

\IEEEdisplaynontitleabstractindextext
 
\IEEEpeerreviewmaketitle

\IEEEraisesectionheading{\section{Introduction}\label{sec:introduction}}
\IEEEPARstart{C}{ontinual} Learning enables a model to continually adapt to new data or tasks, breaking through the traditional one-off training manner in deep learning models. The primary goal of CL is to resolve the stability-plasticity dilemma. Stability refers to the model's need to retain previously learned knowledge, while plasticity emphasizes the model's ability to adjust its weights for new tasks. This challenge is particularly common in real-world applications, such as automated pilot systems and satellite-based remote sensing systems, where data and tasks accumulate over time.  
\begin{figure}[htbp]
	\centering
	\includegraphics[scale=0.61]{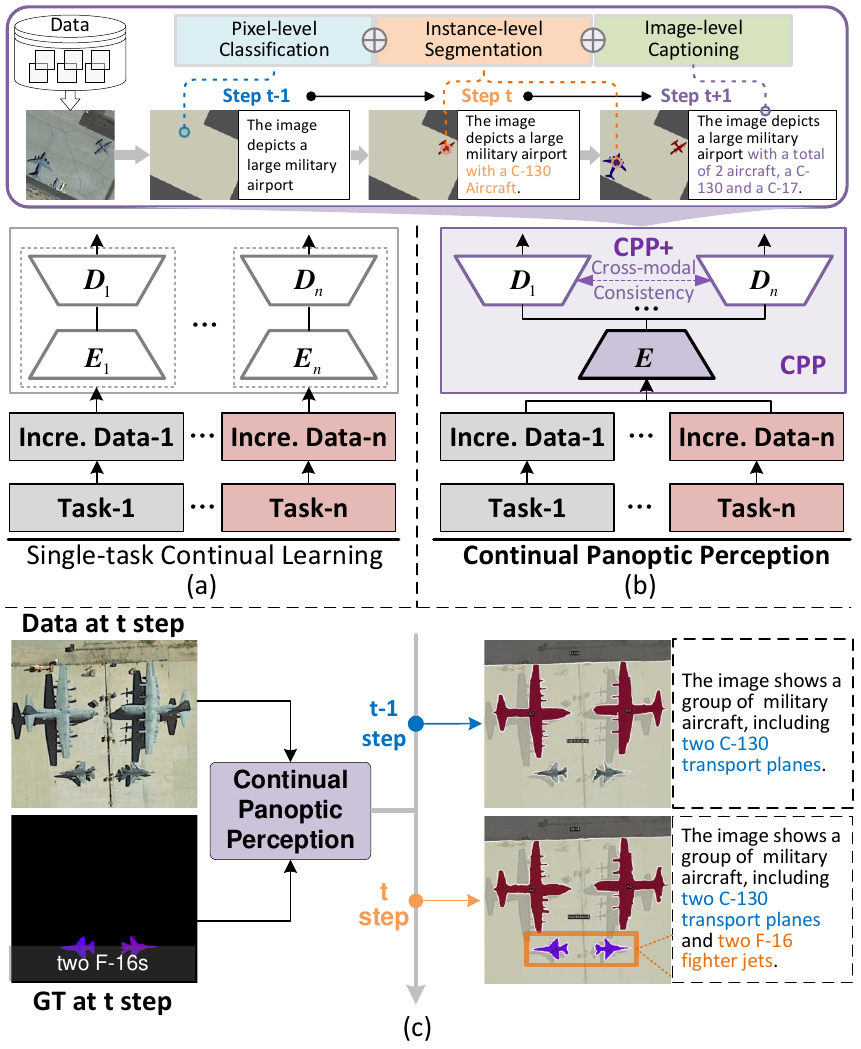}
	\caption{Schematic illustration of the proposed method. (a) Single-task CL methods only support separate training on different tasks. (b)  CPP enables a shared encoder across multimodal tasks, CPP+ integrates multimodal embedding within an end-to-end model. (c)  CPP achieves class-incremental pixel classification, instance segmentation and image captioning.}
	\label{fig:motivation}
\end{figure}

Current CL methods have covered many tasks including image classification~\cite{LWF, masana2022class}, object detection~\cite{Joseph2021IncrementalOD, shmelkov2017incremental}, image segmentation~\cite{IDEC, LAG} and image captioning~\cite{nguyen2019contcap, del2020ratt}, etc.  However, the complex and practical applications urge the model to have the capacity for CL in multi-task learning (MTL). For example, the continuously incremental data in remote sensing usually requires the model to have the ability of continually interpretation on new data, semantics and tasks. Over the past decade, CL has been intensively concerned since it can overcome the typical one-time training schema and enable the model to evolve with continuous data. However, CL encounters two main challenges including catastrophic forgetting and semantic drift~\cite{yuan2023survey}. These problems occur when the parameter updates, resulting in the loss of previously learned knowledge, therefore leading to prediction chaos and model degradation.
Traditionally, the popular fully-supervised methods conduct a complete re-training on the incremental data that may result in an analogous issue, where the model tends to forget its previously learned knowledge due to parameter changes~\cite{LWF}. Besides, current CL approaches face challenges in the trade-off between preserving old knowledge and learning new ones. A range of researches~\cite{RainbowMC, IL2MCI, UsingHT} propose to retrospect known knowledge including sample selection as exemplar memory~\cite{Rolnick2019ExperienceRF, EndoCSS}. However, these replay-based methods usually bring extra memory costs and arise privacy concerns. Another kind that does not rely on old data, instead,  these approaches utilize transfer learning manners like knowledge distillation to inherit the capability of the old model~\cite{EWF, Incrementer}. Nevertheless, current CL approaches focus on single-task incremental learning, while the multimodal and multitask CL lack of exploration and technique pipeline. On the other hand, the complex semantic relations and fine-grained semantic classes make multimodal CL extremely challenging. And there is lack of effective cross-modal collaboration for CL tasks. 

As depicted in Fig.~\ref{fig:motivation}(a), typical single-task CL approaches can only support separate training on different tasks, which limits the CL capacity on complex practical scenarios. In this paper, we propose a CL architecture for panoptic perception~\cite{FineGrip} namely Continual Panoptic Perception (CPP). As illustrated in Fig.~\ref{fig:motivation}(b), CPP facilitates multimodal and multi-task CL within an integrated model that utilizes a shared image encoder for multimodal interpretation. Specifically, CPP comprises a collaborative cross-modal encoder (CCE) that serves as the feature extractor, a malleable continual knowledge distillation (MCKD) module designed for knowledge retention and incremental adaptation, and a self-supervised asymmetric pseudo-labeling (SAPL) mechanism for exemplar-free continual training. Based on CPP, we also propose a cross-modal bidirectional consistency (CBC) constraint which synchronizes multimodal incremental learning, aligning with a stronger architecture referred to as CPP+ to achieve more integrated and robust multimodal CL performance.

Concretely, the CCE module extracts image features with multimodal incremental annotations, which are projected to mask embeddings and text embeddings synchronously for the corresponding multimodal decoder branches. To mitigate catastrophic forgetting, the proposed MCKD module utilizes multimodal contrastive distillation to address semantic drift and reduce semantic chaos. While CBC leverages cross-modal similarity to establish explicit constraint  that enhance the efficiency of multi-task continual learning. Based on the pseudo labeling scheme, the high-confident predictions are transferred to the new step training. SAPL integrates the pseudo labels to improve the label confidence for CL steps. Fig.~\ref{fig:motivation}(c) illustrates the proposed CPP training pattern that synchronously proceeds classification, segmentation and captioning tasks within an integrated model. At each CL step, multimodal new semantics are learned while achieving compatibility with the old knowledge. 

Here we would like to briefly compare the proposed model with general foundational models, of which the latter shows great capacity in zero-shot reasoning and generalization. However, foundational models often require access to large amounts of data, which can raise privacy concerns and storage burdens. The customized CL models show advantages in proprietary knowledge adaptation and low-cost training. Particularly, the proposed CPP model shows multimodal incremental ability with minimal retraining and exemplar-free training approach, which makes it more suitable for applications where proprietary or domain-specific knowledge is critical, such as remote sensing, medical imaging, or industrial automation.

The main contributions are summarized as follows. 
\begin{itemize}
	\item[$\bullet$] We propose continual panoptic perception, a multimodal continual learning method covering pixel-level classification, instance-level segmentation and image-level captioning synchronously. 
	\item[$\bullet$] We develop a cross-modal knowledge distillation scheme,  utilizing cross-task knowledge inheritance and exemplar-free pseudo labeling to reconcile stability and plasticity.
	\item[$\bullet$] A cross-modal embedding consistency constraint is explicitly modeled, harmonizing cross-modal interpretation results with perceptual coherence.
	\item[$\bullet$] The proposed method integrates an end-to-end continual learning framework, supported by extensive experimental validation, which also proves the feasibility of joint optimization across multimodal continual learning tasks.
\end{itemize}

\section{Related Work}
\subsection{Continual Learning} 
Continual learning (CL) originates from as early as~\cite{McCloskey1989CatastrophicII} and has been explored in various fields, including computer vision~\cite{LWF}, natural language processing~\cite{zhang2023vqacl}, remote-sensing~\cite{Marsocci2022ContinualBT}, etc. CL has garnered significant attention in recent years due to its potential to enable models to learn from a continuous stream of data. A key challenge in CL is catastrophic forgetting, a phenomenon that occurs when parameter updates lead to the loss of previously learned knowledge. That is algorithms trained with backpropagation suffer from severe degeneration.This issue is akin to the gradual forgetting experienced by humans when recalling previously learned tasks. To mitigate this problem,  a variety of research efforts have been undertaken, including exemplar replay~\cite{Rolnick2019ExperienceRF, van2020brain}, prototype preserving~\cite{wei2023online, de2021continual},  scalable learning structures~\cite{schwarz2018progress, yoon2019scalable}, knowledge distillation~\cite{li2022learning, wang2024layer}, and meta learning~\cite{javed2019meta, beaulieu2020learning}, etc. Recent advancements have also seen use of  foundation models~\cite{mcdonnell2024ranpac, shi2024continual}, where large-scale pre-trained models leverage transfer learning to adapt their already learned representations to new tasks, significantly improving the efficiency of the learning process.

Despite these advancements, challenges remain, particularly in balancing plasticity and stability to maintain performance across multiple tasks. On the other hand, current CL approaches still focus on single-task incremental learning manner, which limits CL applicability in real-world scenarios where models often encounter multiple tasks simultaneously.

\subsection{Continual Image Segmentation}
Continual image segmentation reforms dense prediction in an incremental training manner. The main challenges are catastrophic forgetting and semantic drift, which arise from the absence of old data and parameter updates~\cite{Hu2021DistillingCE, RW}. According to whether relying on old data~\cite{yuan2023survey}, the CL methods can be divided into replay-based methods~\cite{EndoCSS, ProCA, SSUL, MicroSeg} and exemplar-free methods~\cite{REMINDER, MiB, PLOP, EWF, DKD, ILT, RCIL, IDEC, Incrementer}. The former involves storing a portion of past training data or features as exemplar memory, which brings extra memory costs and raises privacy concerns. The latter usually utilizes knowledge distillation and weight transfer to inherit the capability of the old model. Depending on CL task scenarios, it can be divided into domain-incremental~\cite{BAFFT}, class-incremental~\cite{LAG} and modality-incremental~\cite{CPP} formats, etc. Considering reducing reliance on annotations, few-shot CL methods are also deeply explored in~\cite{SRAA, EHNet, FSCILSS}. Due to the zero-shot ability of large foundation models, there has been growing research focused on enhancing these pre-trained models through weakly supervised incremental training~\cite{FSCILSS}. However, current continual segmentation methods are also mainly designed for single tasks like semantic segmentation~\cite{IDEC, LAG}, the multi-task continual learning has not been deeply studied. 

\subsection{Cross-modal Learning}
In recent years, cross-modal learning has gained significant attention due to its ability to leverage information from multiple modalities. Cross-modal learning has been widely used in various tasks including image segmentation~\cite{ye2019cross, jaritz2020xmuda}, multimodal matching~\cite{lee2018stacked, xu2020cross}, image generation~\cite{kaur2021comparative, zhang2021cross}, etc. One of the fundamental challenges in cross-modal learning is to ensure the alignment of representations across different modalities. For example, CLIP series~\cite{CLIP, MaskCLIP, ZegCLIP} jointly train image and text encoders on large-scale image-text pairs and achieve fair zero-shot performance. Recent advancements also include the integration of attention-based mechanisms~\cite{pan2023fine, ilinykh2022attention}. A prominent example is the visual-linguistic transformers that employ cross-attention layers to interactively fuse information from visual and textual sources, yielding enhanced performance in tasks such as image captioning~\cite{CPP} and visual question answering~\cite{zhao2024see}. For multimodal continual learning task, we propose a cross-modal embedding consistency constraint for multimodal incremental learning.

\subsection{Multi-task Learning}
Multi-task learning (MTL) seeks to jointly learn multiple related prediction tasks with shared information across tasks~\cite{9392366, zhang2018overview, ruder2017overview}. Existing MTL studies primarily focus on feature-based and parameter-based methods. The feature-based approach assumes that different tasks share a common feature representation. In deep learning, it learns a common feature representation for multiple tasks by sharing feature layers in a similar architecture~\cite{sun2020adashare, ding2021mssm}. While parameters-based methods leverage parameter sharing~\cite{rahimian2023dynashare, choi2023dynamic} across tasks, which mainly include hard sharing~\cite{caruana1993multitask} and soft sharing~\cite{yang2023adatask}. The main challenges of MTL include negative optimization and the \textit{seesaw} phenomenon. Although MTL has been explored in incremental learning on classification~\cite{kanakis2020reparameterizing}, object detection~\cite{liu2020multi} and image segmentation~\cite{tian2022multi}, these studies still remains single-task interpretation. However, the multi-objective continual MTL still has a long way ahead. In this paper, we are committed to developing a multi-task continual panoptic perception.

\section{Preliminaries}
The proposed multimodal continual learning task is defined as follows. Considering $\mathcal{D}=\{(x_i, y_i, r_i)\}$ as the training dataset, where $x_i \in \mathbb{R}^{C\times H\times W}$ and $y_i \in \mathbb{Z}^{H\times W}$ denote the training image and mask annotation, respectively. $r_i$ is the captioning annotation. At $t$ step, $\mathcal{D}^t$ indicates the data for incremental training, $C^{0:t-1}$ represents the previously learned classes and $C^t$ is the classes for incremental learning. When training on $\mathcal{D}^t$, the training data of old classes, i.e., $\{\mathcal{D}^0, \mathcal{D}^1, \cdots, \mathcal{D}^{t-1}\}$ is inaccessible. Using $M^{t-1}$ and $M^t$ to represent the \textit{t-1} and \textit{t} step model, respectively. The total training process should consist of \{Step-0, Step-1, $\cdots$, Step-T\} steps.

\section{Method}
The proposed CPP framework forms an integrated framework that includes a collaborative cross-modal encoder (Sec.~\ref{sec:CCE}), a malleable continual knowledge distillation scheme (Sec.~\ref{sec:MCKD}), a cross-modal bidirectional consistency constraint (Sec.~\ref{sec:CBC}) and a self-supervised asymmetric pseudo-labeling method (Sec.~\ref{sec:SAPL}).

\subsection{Collaborative Cross-modal Encoder}
\label{sec:CCE}
In the proposed architecture, a shared encoder is designed for multi-task interpretation including dense prediction and image captioning. The former can be seen as a segmentation task and the latter is a global semantic understanding task. Thus we utilize a unified Transformer architecture to cope with cross-modal feature extraction. 
\begin{figure*}[tb]
	\centering
	\includegraphics[scale=0.63]{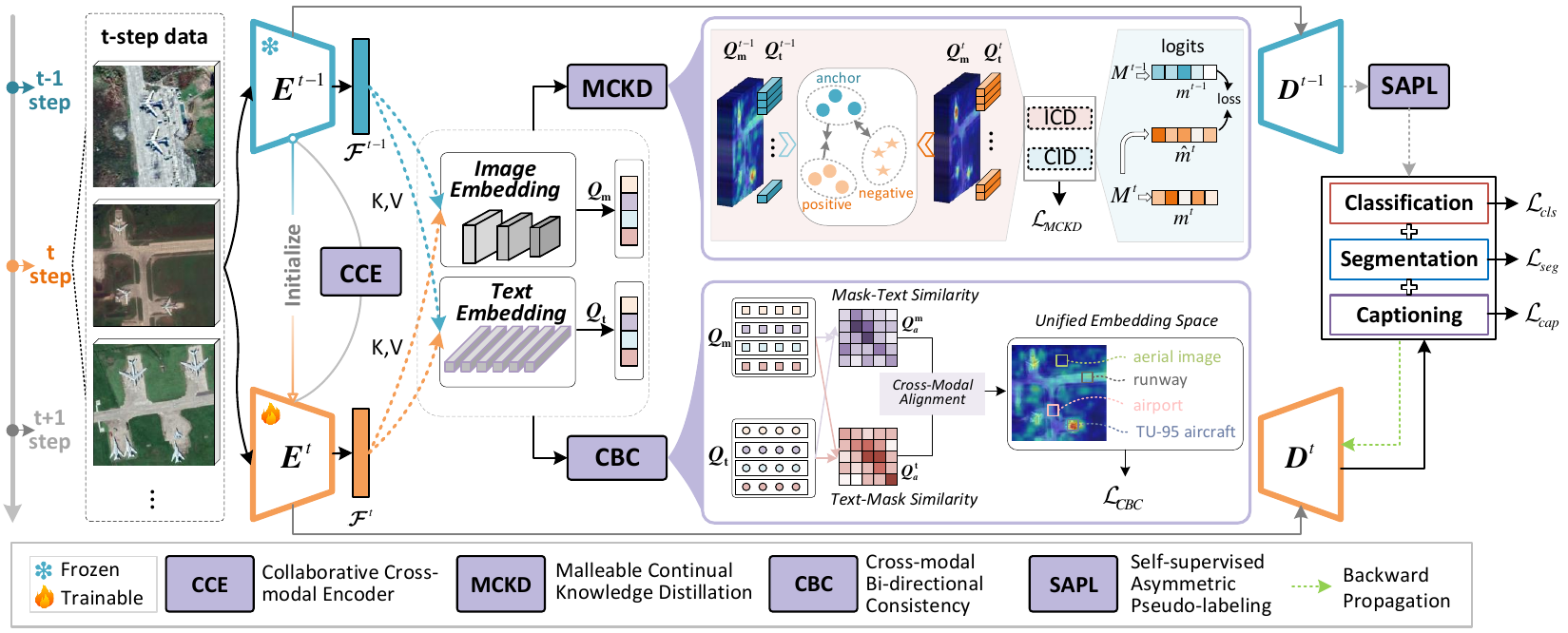} 
	\caption{The framework of the proposed method. The input consists of the incremental images with corresponding multimodal annotation. The output consists of the mask predictions for both old and new classes and image captioning result with new semantics}
	\label{fig:network}
\end{figure*}  

As illustrated in Fig.~\ref{fig:network}, considering an image $x\in \mathbb{R}^{C\times H\times W}$, we firstly apply a model-agnostic feature extractor to extract features. Concretely, the image encoder $E$ produces a downsampled feature $\mathcal{F}\in \mathbb{R}^{N\times H'\times W'}$. The image feature $\mathcal{F}$ derived from CCE can be utilized for both mask prediction and caption generation. Concretely, the encoded image feature from CCE is:
\begin{equation}
	\mathcal{F} = \Theta(ReLU(\Theta(E(x))))+E(x)
\end{equation}
where $E(\cdot)$ is the encoding process. $\Theta(\cdot)$ is a linear mapping.

Whereafter the decoder is decoupled to a Transformer for mask prediction, one-to-one mappings between the mask predictions and the ground truth are generated.  And a Transformer for word prediction. For pixel-level predication, the $\mathcal{F}$ is firstly generated to $N$ mask embeddings via a Transformer decoder with $N$ query input.  The decoder then upsamples $\mathcal{F}$  to obtain per-pixel feature embedding $\mathcal{E}\in \mathbb{R}^{C_e\times H \times W}$.
We regard both instance and semantic segmentation as mask classification problems and handle them with a Transformer-based architecture. Concretely, assuming \textit{N} learnable queries $Q\in \mathbb{R}^{C_q\times N}$, $\mathcal{F}$ is used as keys (K) and value (V). A standard Transformer decoder is used to update $Q$. In the mask classification branch, the encoded query undergoes a linear transformation to yield a classification result of $N \times (C +1)$, where $C$ is the total number of foreground and background categories. To make full use of shared features, the query is projected into mask embeddings $\textbf{Q}_m\in \mathbb{R}^{N\times C_e}$ in the mask generation branch, which has the same channel dimensions as the per-pixel feature embeddings. And the text embedding $\textbf{Q}_t$ for captioning. 

Specifically for the captioning task, a standard Transformer decoder consisting of multiple decoder layers is used for sentence prediction, each having a masked self-attention mechanism, a cross-attention mechanism, and a feed-forward neural network. 
The self-attention mechanism can tackle the long-range context, which is beneficial to both segmentation and captioning tasks.

\subsection{Malleable Continual Knowledge Distillation}
\label{sec:MCKD}
To inherit the capacity from the old model in the absence of old data, we propose a malleable continual knowledge distillation (MCKD) method under the circumstances of only the previous model $M^{t-1}$ and the incremental data $\mathcal{D}^{t}$ can be accessed.  For fine-grained CL tasks, it faces misclassification challenges while the old classes and future classes are mixed in \textit{bg} class during CL steps. There are proofs showing the latent domain gap between the cross-modal data could lead to a learning ambiguity during knowledge distillation~\cite{ren2021learning}. Here we propose a cross-task contrastive distillation method based on task-interactive guidance. 

Firstly, using $\textbf{Q}_{m}^{t-1}$ and $\textbf{Q}_{m}^{t}$ indicate the mask embeddings, while $\textbf{Q}_t^{t-1}$ and $\textbf{Q}_{t}^{t}$ for text embeddings from $t-1$ and $t$ steps, respectively.  	 
Concretely, the MCKD module consists of Intermediate Contrastive Distillation (ICD) and Cross-guided Instance Distillation (CID).

\noindent\textbf{Intermediate Contrastive Distillation}. Since the segmentation and the captioning embeddings are both derived from the same features $\mathcal{F}$ produced by CCE, the ICD is conducted on both tasks. To alleviate the classifier confusion caused by semantic drift, we propose a contrastive distillation across $\textbf{Q}_{m}$ and $\textbf{Q}_{t}$. Inspired by~\cite{IDEC}, we argue that using embedding from the old model is more confident to reduce prediction error since catastrophic forgetting. Concretely, the optimization objective contains global output logits and class-wise contrastive learning between old and new classes. Thus the constraint for ICD is:
\begin{equation}
	\mathcal{L}_{ICD} = d(\textbf{Q}_m^{t-1}, \textbf{Q}_m^{t}) + d(\textbf{Q}_t^{t-1}, \textbf{Q}_t^t) + D_{CL}(f_a^{t-1}, f_p^{t}, f_n^{t})
\end{equation}
\begin{equation}
	D_{CL} = \frac{1}{|C^{0:t}|}\sum_{i, j,  i\neq j}^{C^{0:t}} \sum_k^{C^{0:t-1}} \mathbbm{1}[i=k][d(f_k^{t-1}, f_i^{t})-d(f_k^{t-1}, f_j^{t})]
\end{equation}
where $d(\cdot)$ is a similarity measure to constraint the representation consistency between $M^{t-1}$ and $M^{t}$. $f_a^{t-1}$, $f_p^{t}$ and $f_n^{t}$ represent the \textit{anchor}, \textit{positive} and \textit{negative} embeddings, respectively. $f_k^{t-1}$ indicates the embedding belonging to \textit{k}-th class from $M^{t1}$. $f_i^{t}$ and $f_j^{t}$ indicate the corresponding embeddings from $M^t$.

\noindent\textbf{Cross-guided Instance Distillation}. In fine-grained interpretation tasks, the large amount of instances and complex distribution increase the probability of semantic fusion. Thus the instance segmentation requires fine-grained distillation to avoid catastrophic forgetting. As many pioneers used for CL tasks, directly distilling the feature map without considering foreground regions would result in training with a large amount of background information, leading to insufficient learning of important foreground regions and poor distillation performance. Here we leverage a prior of credibility that the prediction from old model is more confident than that from the new model since catastrophic forgetting in CL tasks. Thus we propose a distillation method emphasizing the guidance from the old model. Since Chen et al.~\cite{chen2022bevdistill} propose a cross-modal instance distillation, we adapt it to multi-task CL scenarios. The quality score $q_r$ serves as an indicator to guide the student on which teacher's predictions should be paid more weight.
\begin{equation}
	q_r = (c_r)^{\gamma}\times IoU(m_r^{t-1}, \hat{m}_r^t)^{(1-\gamma)}
\end{equation}
where $\gamma$ indicates the weight of classification and segmentation. $c_r$ indicates the predictive classification results of instance $r$. $m^{t-1}$ and $\hat{m}^{t}$ are the segmentation results of $M^{t-1}$ and $M^{t}$, respectively.  It is noted that $\hat{m}_r^{t}$ is the student output from the teacher decoder to restrain the model inheritance~\cite{CrossKD}. For fine-grained tasks, the recognition branch should be assigned a higher weight to alleviate misclassification due to large intra-class variance. Thus a constraint between $M^{t-1}$ and $M^t$ is proposed to optimize the instance segmentation task via a guided instance distillation loss. 

The CID objective is defined as:
\begin{equation}
	\mathcal{L}_{CID} = \sum_i[-q_r(d(y_r^{t-1}, y_r^{t}))]
\end{equation}
where $y_r$ indicates the predicted classification result of instance $r$. Thus the MCKD objective is the combination of ICD and CID:
\begin{equation}
	\mathcal{L}_{MCKD} = \mathcal{L}_{ICD} + \mathcal{L}_{CID}
\end{equation}

\begin{figure}[tb]
	\centering
	\includegraphics[scale=0.6]{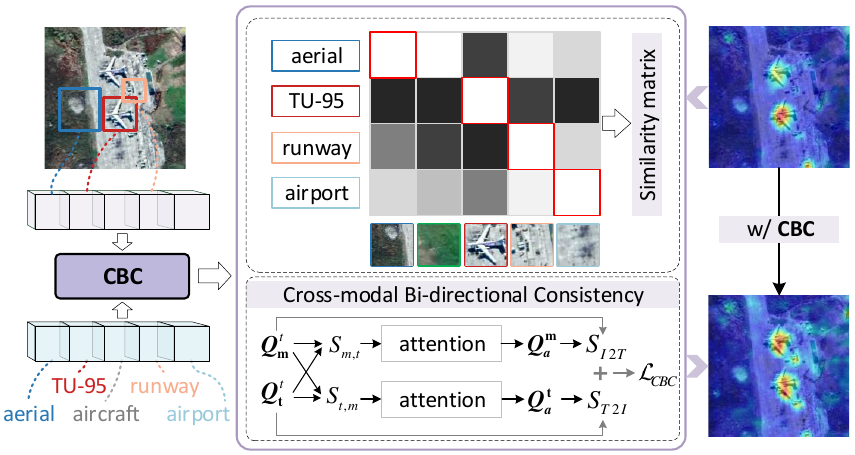} 
	\caption{Cross-modal bi-directional consistency constraint.}
	\label{fig:CBC}
\end{figure}  

\subsection{Cross-modal Bidirectional Consistency Constraint}
\label{sec:CBC}
Since the image and text embedding output from different decoders tend to be inconsistency due to the independence of each modal. However, it tends to aggravate the semantic chaos during CL steps, leading to model degradation and prediction collapse. 

Based on the cross-modal embeddings from CCE, we propose a consistency constraint for panoptic perception task. As depicted in Fig.~\ref{fig:CBC}, $\textbf{Q}_m$ and $\textbf{Q}_{t}$ are first projected to the unified embedding space, to achieve dimensional consistency. Firstly, the cosine similarity is computed between the image and text embeddings from $t-1$ step and $t$ steps based on the incremental data, which manifests as the mask-text pairs. Particularly, a fully connect layer is added to transform $\mathcal{F}$  to a embedding vector \textbf{Q}. Then the cosine similarity between image and text embeddings can be formulated as follows. 
\begin{equation}
	S_{m,t} = \frac{\textbf{Q}_m^{t} \cdot \textbf{Q}_t^{t}}{\Vert \textbf{Q}_m^{t} \Vert \Vert \textbf{Q}_t^{t} \Vert}
\end{equation}
where $\Vert \cdot \Vert$ is the L2 norm. 

Inspired by~\cite{xu2020cross}, we build a Cross-modal Bidirectional Consistency (CBC) constraint across image and text embeddings from an interactive manner to align the cross-modal information. For image to text ($I \rightarrow T$) fusion, the optimization direction should be maximize the similarity between the incremental $\textbf{Q}_{m}^{t}$ and $\textbf{Q}_{t}^{t}$. We define the weighted attention as $\textbf{Q}_{a}^m$.
 
\begin{equation}
	\begin{aligned}
		\textbf{Q}_{a}^m &= \sum \alpha_{1,i} \textbf{Q}_{t,i}^{t}, i\in [0,N-1], \\ &s.t., \alpha_1 = \frac{\exp(\eta_1 S_{m,t})}{\sum_i \exp(\eta_1 S_{m,t,i})}
	\end{aligned}
\end{equation}
where $N$ indicates the learned class number. And the softmax function was used to normalize the similarity score as $\alpha_1$. $\eta_1$ is used as temperature parameter controls the sharpness of the weight distribution. 
Thus the image to text similarity fusion can be formulated as follows.
\begin{equation}
	S_{I2T} = \frac{ \textbf{Q}_m^t \cdot \textbf{Q}_a^m}{\Vert \textbf{Q}_m^t \Vert \Vert \textbf{Q}_a^m \Vert}
\end{equation}

Likewise, for text to image ($T \rightarrow I$) fusion, the similarity can be formulated as $S_{t,m}$:
\begin{equation}
	S_{t,m} = \frac{\textbf{Q}_t^{t} \cdot \textbf{Q}_m^{t}}{\Vert \textbf{Q}_t^{t} \Vert \Vert \textbf{Q}_m^{t} \Vert}
\end{equation}

Then the weighted attention $\textbf{Q}_a^t = \sum \alpha_2 \textbf{Q}_{m,i}^t$, where $\alpha_2= \frac{\exp(\eta_2 S_{t,m})}{\sum_{i} \exp(\eta_2 S_{t,m,i})}$ indicates the normalized attention weight. Thus the similarity is computed as follows.

\begin{equation}
	S_{T2I} = \frac{ \textbf{Q}_t^t \cdot \textbf{Q}_a^t}{\Vert \textbf{Q}_t^t \Vert \Vert \textbf{Q}_a^t \Vert}
\end{equation}

As depicted in Fig.~\ref{fig:CBC}, the cross-modal alignment mechanism dynamically adjusts the attention weights to accommodate the data distribution of new semantics. This helps to alleviate semantic chaos and encourages the model to effectively align with fine-grained semantics.
Concretely, to align image and text modal information, the proposed CBC is to minimize the similarity cost via bidirectional joint optimization:
\begin{equation}
	\mathcal{L}_{CBC} = (1-S_{I2T}) + (1-S_{T2I})
\end{equation}

\subsection{Self-supervised Asymmetric Pseudo-labeling}
\label{sec:SAPL}
\begin{figure}[tb]
	\centering
	\includegraphics[scale=0.75]{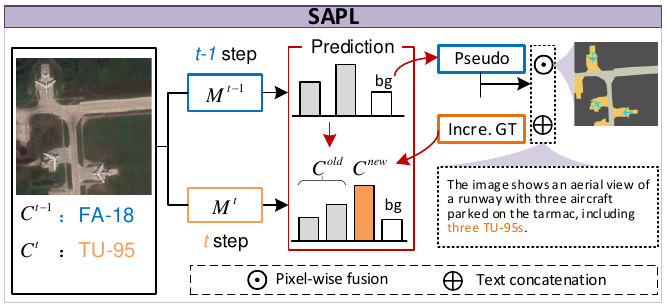} 
	\caption{Self-supervised pseudo-labeling. The asymmetric task reliance indicates the pseudo labels are cross-verified by more reliable predictions from multimodal branches.}
	\label{fig:pseudo}
\end{figure}

Considering only the incremental classes are labeled, we propose an exemplar-free pseudo-labeling method. Since predictions from different modals possess different confidences, the pseudo labels are generated with asymmetric task reliance, i.e., more emphasis on confident predictions. As seen in Fig.~\ref{fig:pseudo}, we observe the captioning task focuses more on the global context. While the segmentation tasks tend to cause pixel misclassification within a single intact target area. Using  $\mathcal{Y}_{cap}^{t-1}$ indicates captioning output from the $M^{t-1}$. Thus the segmentation annotation $\bar{y}_i^t$ at $t$ step for pixel $i$ is defined as:
\begin{equation}
	\begin{aligned}
		\bar{y}_i^t = \left\{ 
		\begin{aligned} 
			&\tilde{y}_i^{t-1},  \rm{if} \ (\tilde{y}_i^{t-1} \in C^\text{{0:t-1}}) \land [(p_i \geq \Gamma) \lor  [(\tilde{y}_i^{t-1}\in \mathcal{Y}_{cap}^\text{t-1})]\\
			&y_i^t, \quad \rm{if} \ (x_i\in C^t) \\ 
			&c^b, \quad otherwise \\
		\end{aligned}
		\right. \\
	\end{aligned} 
\end{equation}
where $\tilde{y}_i^{t-1}$ is the pseudo label for pixel $i$ generated from $M^{t-1}$, $p_i$ is the predictive probability of pixel $i$ at $t$ step. $x_i$ is the $i$-th pixel in image $x$. $y_i^t$ is the incremental annotation for pixel $i$ at $t$ step. $c^b$ indicates the unknown background class. The asymmetry manifests itself by threshold or more rely on captioning predictions.
Since the learned and future classes are mixed in $c^b$ at each CL step, the relative scoring of softmax could lead to significant semantic chaos and catastrophic forgetting during CL training steps. Instead, we take independent \textit{Sigmoid} with binary cross-entropy loss to cope with the variable class number during CL steps and alleviate the interference from variable background semantics. 

For the captioning task, every sentence starts with ``START" and ends with ``END" keywords. The label at $t$ step is concatenated with the pseudo label before "END". Thus the captioning label for $t$ step is:
\begin{equation}
	\bar{y}_{cap}^t = \tilde{y}_{cap}^{t-1} \oplus {y}_{cap}^{t} 
\end{equation}
where $\tilde{y}_{cap}^{t-1}$ indicates the predicted captioning from $M^{t-1}$ and ${y}_{cap}^{t} $ is the readily available label at $t$ step. $\oplus$ indicates the concatenation operation.

\subsection{Overall Objective via MTL}
The proposed model conducts end-to-end multimodal training with weighted   losses across multi-objective joint training. Concretely, the objective considers five parts: classification loss $\mathcal{L}_{cls}$, segmentation loss $\mathcal{L}_{seg}$,  caption generation loss $\mathcal{L}_{cap}$, knowledge distillation loss $\mathcal{L}_{MCKD}$ and cross-modal consistency loss $\mathcal{L}_{CBC}$. Considering joint training among diverse tasks, we propose a weighted loss for MTL issue in a simple but effective way. Concretely, $\mathcal{L}_{cls}$ is formed by a cross-entropy loss.  And $\mathcal{L}_{seg}$ is binary mask loss:
\begin{equation}
	\mathcal{L}_{seg} =\sum_{i=1}^N [\mathds{1}[c_i^{gt} \ne \emptyset]\mathcal{L}_{m}(\hat{y}, \bar{y})]
\end{equation}
where $\mathcal{L}_{m}$ is formed by focal loss~\cite{focalloss} and dice loss~\cite{diceloss}:
\begin{equation}
	\mathcal{L}_{m} =  \eta_1 \mathcal{L}_{focal}(\hat{y}, \bar{y}) + \eta_2 \mathcal{L}_{dice}(\hat{y}, \bar{y}) 
\end{equation}
where $\eta_1$ and $\eta_2$ are the weighting component.  $c_i^{gt}$ is the label for class $c$. And the captioning loss is formed by a cross-entropy loss:
\begin{equation}
	\mathcal{L}_{cap} = -\sum_{t=1}^{L}log p_{cap}(\bar{y}_{cap})
\end{equation}
where $p_{cap}$ is the word prediction probability from the caption module and $\bar{y}_{cap}$ represents the annotation. Thus the integrated objective is defined as:
\begin{equation}
	\mathcal{L}=\mathcal{L}_{cls}+\mathcal{L}_{seg}+\lambda \mathcal{L}_{cap} +  \mathcal{L}_{MCKD} + \mathcal{L}_{CBC}
	\label{eqn-overall}
\end{equation}
where $\lambda$ is the weight of captioning tasks.  Specifically, the distillation loss is only applied at CL steps.

\section{Experiment}
\subsection{Datasets and Protocols}
\textbf{Datasets}:  We choose two fine-grained perception datasets including FineGrip~\cite{FineGrip} and ADE20K~\cite{ADE} to validate the proposed model.
\begin{itemize}
	\item[$\bullet$] FineGrip~\cite{FineGrip} is a multi-task remote-sensing dataset that includes 2649 images, with 12054 fine-grained instance segmentation masks belonging to 20 foreground classes (things), 7599 background (stuff) semantic masks covering 5 classes, and 13245 fine-grained sentence description annotations. The sample number in the training set is much smaller than that in the validation set, which provides challenging but practical scenes for CL under limited data conditions. Note that all \textit{stuff} classes ($C^{21-25}$) are trained at the initial step since they are covered in most samples. We evaluate our model on 20-5 (2 steps), 15-5 (3 steps), 15-2 (6 steps) and 10-5 (4 steps) tasks, respectively.
	
	\item[$\bullet$] ADE20K~\cite{ADE} is a large-scale semantic segmentation dataset containing 150 classes that cover indoor and outdoor scenes. The dataset is split into 20210 training images and 2000 validation images. It covers a wider variety of classes in natural scenes with abundant instances and categories. We evaluate the model on 100-50 (2 steps) and 100-10 (6 steps) settings. 
	
	\item[$\bullet$] COCO~\cite{COCO} contains segmentation and caption annotations.  However, the segmentation annotations and captions do not exhibit fine-grained semantic alignment. Consequently, we utilize this dataset to validate the model's capability to leverage cross-modal information for CL. Specifically, two CL tasks including 40-40 (2 steps) and 40-10 (5 steps) are conducted to verify the model. 
\end{itemize}

\textbf{Protocols}: There are mainly two different CL settings: \emph{disjoint} and \emph{overlapped}. In both settings, only the current classes $C^t$ are labeled and others are set as background. In the former, images at $t$ step only contain $C^{0:t-1} \cup C^t \cup C^{bg}$. While the latter contains $C^{0:t-1} \cup C^{t} \cup C^{t+1: T} \cup C^{bg}$, which is more realistic and challenging. In this study, we focus on \emph{overlapped} setting. We also report two baselines, i.e., \emph{fine-tuning} on $C^{t}$,  and training on all classes \emph{offline}. The former is the lower bound and the latter can be regarded as the upper bound of this task.

\textbf{Metrics}: We compute the panoptic quality (PQ), segmentation quality (SQ) and recognition quality (RQ) to measure the segmentation efficiency. While the Bilingual Evaluation Understudy (BLEU)~\cite{papineni2002bleu} is used to evaluate the generated text. The metrics is computed as follows.
\begin{equation}
	\text{PQ} = \frac{\sum_{(p,g) \in \text{TP}} \text{IoU}(p, g)}{|\text{TP}| + \frac{1}{2}|\text{FP}| + \frac{1}{2}|\text{FN}|}
\end{equation}
where $\text{TP}$, $\text{FP}$, and $\text{FN}$ represent true positives, false positives, and false negatives, respectively. A pair $(p,g) \in \text{TP}$ denotes a match between the predicted mask and the ground truth. Further, the PQ can be decomposed into Segmentation Quality (SQ) and Recognition Quality (RQ) to separately measure the performance of segmentation and recognition:
\begin{equation}
	\text{SQ} = \frac{\sum_{(p,g) \in \text{TP}} \text{IoU}(p, g)}{|\text{TP}|}
\end{equation}
\begin{equation}
	\text{RQ} = \frac{|\text{TP}|}{|\text{TP}| + \frac{1}{2}|\text{FP}| + \frac{1}{2}|\text{FN}|}
\end{equation}

The BLEU score is defined as:
\begin{equation}
	\text{BLEU} = BP \cdot \exp \left( \sum_{n=1}^{4} w_n \log p_n \right)
\end{equation}
where:
\begin{equation}
	p_n = \frac{\text{Count of matching n-grams}}{\text{Count of total n-grams in generated text}}
\end{equation}

$w_n$ is the weight for n-grams, typically set to $\frac{1}{4}$ for BLEU-4.

The brevity penalty, \( BP \), is defined as:
\begin{equation}
	BP = 
	\begin{cases} 
		1 & \text{if } c > r \\
		\exp(1 - \frac{r}{c}) & \text{if } c \leq r 
	\end{cases}
\end{equation}
where $c$ is the length of the generated translation and $r$ is the effective reference corpus length.

For continual semantic segmentation task, we compute mean intersection-over-union (mIoU) for evaluation: 
\begin{equation}
	mIoU = \frac{TP}{TP+FP+FN}
\end{equation}
where TP, FP and FN are the numbers of true positive, false positive and false negative pixels, respectively.

\subsection{Implementation Details}
We use MaskFormer~\cite{MaskFormer} and Mask2Former~\cite{Mask2Former}  with ResNet-101~\cite{resnet} as the base encoders to extract image features. For all experiments, the initial learning rate is 0.01 and decayed by a \textit{poly} policy. The implementation is based on PyTorch 2.1 with CUDA 12.7 and all experiments are conducted on a workstation with two NVIDIA A800 GPUs. For all CL steps, the training epoch is set to 90. The hyper-parameters are set as $\eta_1=20.0$ and $\eta_2=1.0$. The implementation is based on MMDetection~\cite{mmdetection}. 

\subsection{Quantitative Evaluation}
\subsubsection{FineGrip}
To comprehensively evaluate the CPP performance on old and new classes, we compute the segmentation performance and captioning performance at the initial step and the final step, respectively. The proposed model is tested on three aspects following~\cite{yuan2023survey}:  few-step with multi-class (FSMC), multi-step with few-class (MSFC) and multi-step with multi-class (MSMC). Particularly, FSMC emphasizes the ability to learn new knowledge (plasticity) since a plenty of new classes are adapted in a single step. In contrast, MSFC underlines the ability of anti-forgetting on old knowledge (stability) because many CL steps are conducted. While MSMC provides comprehensive evaluation. The proposed model is evaluated from multimodal CL performance and cross-task CL synergy, which are the key problems beyond the pioneer single-task CL approaches. Besides, we also extend the single-modal models including MaskFormer~\cite{MaskFormer} and Mask2Former~\cite{Mask2Former} for continual segmentation, SAT~\cite{SAT} and VitCAP~\cite{vitcap} for continual captioning.
\begin{table*}[t]
	\caption{Quantitative performance on FineGrip dataset. We evaluate the segmentation task with PQ (\%). $C^{o}$, $C^{n}$ and $C^{a}$ are the performance for old classes, new classes and all classes after all CL steps, respectively. The captioning performance is evaluated by BLEU scores (beamsize=5), which are reported before (B$^{b}$) and after (B$^{a}$) all CL steps. $\dagger$ indicates we re-implement the method to CL tasks. \textit{fine-tuning} and \textit{fine-tuning+}  indicate re-training the CPP and CPP+ model only with incremental data. } 
	\label{tab:finegrip}
	\centering
	\setlength{\tabcolsep}{0.2mm}{
		{\begin{tabular*}{0.95\textwidth}{@{\extracolsep{\fill}}l||ccc|cc||ccc|cc||ccc|cc||ccc|cc@{}}
				\toprule[0.5mm]
				\multirow{2}*{Method}&\multicolumn{5}{c}{20-5 (2 steps)}&\multicolumn{5}{c}{15-5 (3 steps)}&\multicolumn{5}{c}{15-2 (6 steps)} &\multicolumn{5}{c}{10-5 (4 steps)}\\
				&$C^{o}$&$C^{n}$&$C^{a}$&$B^{b}$&$B^{a}$  &$C^{o}$&$C^{n}$&$C^{a}$&$B^{b}$&$B^{a}$ &$C^{o}$&$C^{n}$&$C^{a}$&$B^{b}$&$B^{a}$ &$C^{o}$&$C^{n}$&$C^{a}$&$B^{b}$&$B^{a}$ \\
				\midrule
				\textit{fine-tuning}&1.01&4.22&1.65&9.85&4.28&2.85&5.11&3.75&11.42&3.65&1.28&0.77&1.08&11.42&2.22&0.67&0.49&0.56&11.42&1.96\\
				\textit{fine-tuning}+&2.05&5.13&2.67&10.91&6.15&3.73&5.46&4.42&10.17&4.66&1.60&1.00&1.36&10.17&4.73&1.77&1.70&1.73&12.15&3.74\\
				\midrule
				MaskFormer$^{\dagger}$~\cite{MaskFormer}&23.55&31.14&25.07&-&-&25.68&23.01&24.61&-&-&15.17&7.79&12.22&-&- &16.29&12.73&14.15&-&- \\
				Mask2Former$^{\dagger}$~\cite{Mask2Former} &28.59&29.81&28.83&-&-&29.17&24.19&27.18&-&-&20.97&16.35&19.12&-&-&21.39&12.42&16.01&-&- \\
				\midrule
				SAT$^{\dagger}$~\cite{SAT}&-&-&-&21.45&17.41&-&-&-&18.94&16.50&-&-&-&18.94&10.47&-&-&-&14.25&15.13\\
				VitCAP$^{\dagger}$~\cite{vitcap} &-&-&-&22.32&20.53&-&-&-&20.37&18.90&-&-&-&20.37&17.95&-&-&-&22.19&18.55 \\
				\midrule
				\textbf{CPP} &27.50	&\textbf{33.11}	&28.62&35.55&34.12&30.17	&25.84 &28.44 &33.01&27.00 &17.73	&10.62&	14.89&33.01&21.52 &18.55&14.79&16.29&24.93&25.59\\
				\textbf{CPP+} &\textbf{37.61}&32.36&\textbf{36.56}&\textbf{39.31}&\textbf{35.00}  &\textbf{33.64}&\textbf{31.51}&\textbf{32.79}&\textbf{38.63}&\textbf{36.32}  &\textbf{25.70}&\textbf{21.26}&\textbf{23.92}&\textbf{38.63}&\textbf{28.34}  &\textbf{27.60}&\textbf{16.59}&\textbf{20.99}&\textbf{33.36}&\textbf{32.37}\\
				\midrule	
				\textit{offline}&57.30&53.27&56.49 &41.53&41.53&58.85&52.95&56.49&41.53&41.53&58.85 &52.95&56.49 &41.53&41.53&55.20 &57.35&56.49&41.53&41.53 \\
				\bottomrule[0.5mm]
		\end{tabular*}}{}}
\end{table*}
\begin{figure*}[htbp]
	\centering
	\includegraphics[scale=0.23]{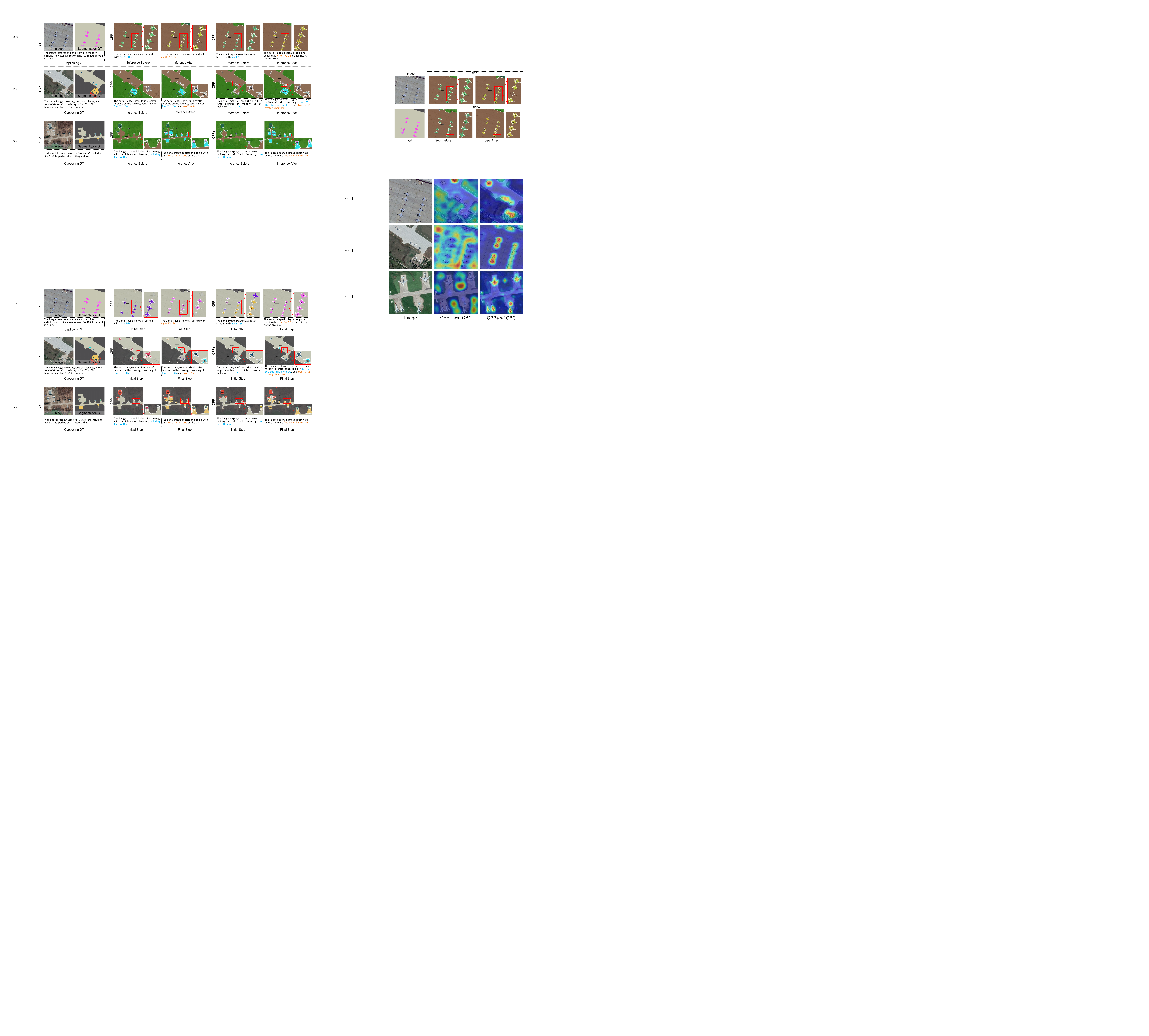} 
	\caption{Qualitative visualization \textit{before} and \textit{after} CL steps. Specifically, the predictions at the \textit{initial step} and the \textit{final step} are displayed. }
	\label{fig-vis-step}
\end{figure*} 

\textbf{Multimodal CL performance}.
As seen in Table~\ref{tab:finegrip}, we first evaluate the PQ for old, new and all classes after all CL steps, respectively. 
Compared to naive fine-tuning approaches, the proposed CPP method achieves superior anti-forgetting on old classes and plasticity on new classes simultaneously. The captioning performance is evaluated from two aspects: The performance at the initial step and after all CL steps. The former is the initial learned result with limited partial classes. While the latter indicates the matching accuracy of the predicted words after all CL steps. On the one hand, the captioning capacity improves since the new semantics are learned to meet with an intact understanding of the image. On the other hand, we observe a measurable decline in BLEU scores throughout the CL process, which can be attributed to the persistent challenge of catastrophic forgetting in continual learning scenarios.

The results in Table~\ref{tab:finegrip} also prove that the joint multi-task optimization in CL tasks should boost sub-task capacity. Compared to the segmentation method MaskFormer under the proposed CL setting, the proposed CPP achieves 3.55\%, 3.83\%, 2.67\% and 2.14\% PQ improvements on all classes on 20-5, 15-5, 15-2 and 10-5 tasks, respectively. In addition, CPP achieves overall improvements across multimodal tasks. Concretely, for 15-2 task, the long-step learning brings severe catastrophic forgetting. As a comparison, CPP maintains relative higher performance on both segmentation task and captioning task, which demonstrates the effectiveness of the proposed multimodal continual learning architecture.

Furthermore, the enhanced CPP+ demonstrates substantial superiority across both segmentation and captioning metrics. It proves the cross-modal consistency strategy with a stronger baseline can boost multimodal CL performance. Particularly in the challenging 15-2 task, CPP+ surpasses CPP by 9.03\% PQ and 6.82\% BLEU, suggesting considerable potential for further advancements in multimodal continual learning systems..

\textbf{Qualitative visualization}.
Fig.~\ref{fig-vis-step} depicts a comprehensive visual comparison of the segmentation and captioning results obtained by CPP and CPP+ across various tasks. It can be informed from two aspects. On the one hand,  the visualization highlights the phenomenon of semantic confusion, where instances are correctly localized but erroneously classified. As seen in the first row in Fig.~\ref{fig-vis-step} before CL training,  the foreground instances are located but misclassified to false class labels.
On the other hand, the unknown semantics can be ignored at the previous steps, as illustrated in the second row in Fig.~\ref{fig-vis-step}. While after the CL training steps by CPP, the semantic chaos is alleviated by distinguishing old and new classes. Meanwhile the captioning results are also replenished by taking into account both new and old semantics. Compared to CPP, CPP+ achieves stronger anti-forgetting with clearer semantic boundaries and precise semantic resolution in captioning generation, which also aligns the results from Table~\ref{tab:finegrip}.

\textbf{Cross-task CL synergy}.
In the proposed multimodal CL architecture, we argue that mutual guidance from different branches in knowledge distillation and pseudo-labeling can improve the performance of each sub-task. To reveal the mutual impact between multimodal branches, we perform separate training on the segmentation task and captioning task, respectively. As shown in Table~\ref{tab:finegrip}, specifically for CPP 15-5, the joint training by CPP achieves 4.49\% and 2.83\% PQ improvement on $C^o$ and $C^n$ compared to the baseline~\cite{MaskFormer} on the old and all classes after CL steps. It also achieves 10.5\% BLEU score higher than the captioning only method~\cite{SAT} after CL training steps. 
On the other hand, CPP+ utilizes cross-modal consistency constraint with a stronger base model, resulting to 7.73\% and 5.61\% PQ improvement against with~\cite{Mask2Former} on 20-5 and 15-5 tasks, respectively. In the challenging 15-2 and 10-5 tasks, CPP+ also achieves 9.03\% and 4.70\% PQ superiority than that of CPP. For captioning task, CPP+ achieves 9.32\%, 6.82\% and 6.78\% BLEU score than that of CPP after all incremental steps on 15-5, 15-2 and 10-5 tasks, showing superiority especially in long-step tasks. The results prove the mutual boosting of joint training across multi-modal tasks in CL problems, which we believe is also the key optimization guidance in addressing multimodal CL tasks.

\subsubsection{ADE20K} 
Similarly, we also evaluated the model under FSMC and MSMC scenarios against with the pioneer continual semantic segmentation models including~\cite{MiB, PLOP, RCIL, IDEC, LAG}.  Particularly, multi-class fine-grained segmentation performance is the main concern on this dataset. Specifically, we evaluate the mIoU since there is only segmentation annotations provided by the dataset. Note that when training on ADE, the CBC module described in Sec.~\ref{sec:CBC} is not involved. The results in Table~\ref{table-ADE} show the segmentation performance evaluated by mIoU. 

\begin{figure}[htbp]
	\centering
	\includegraphics[scale=0.42]{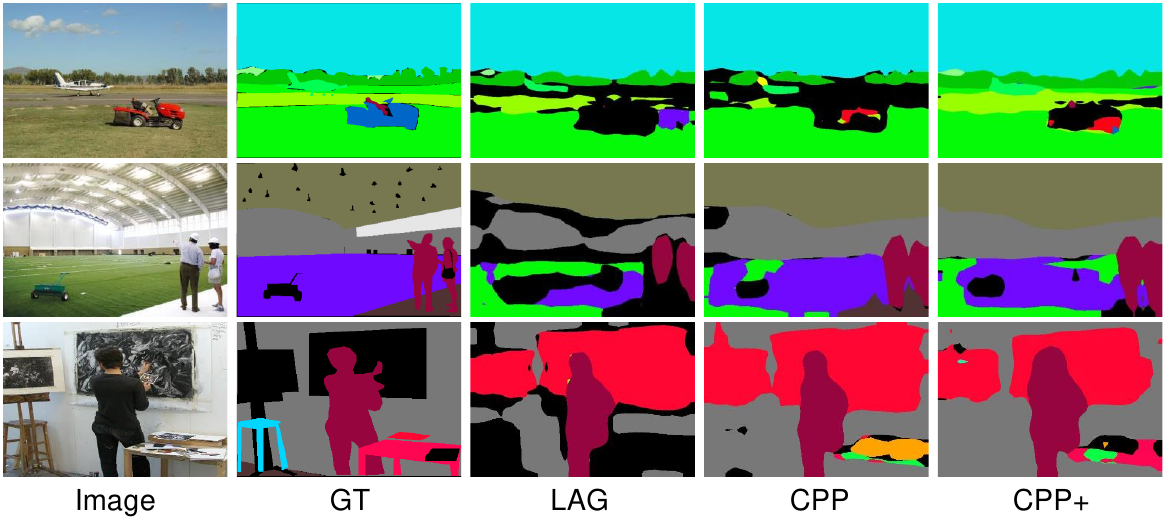}
	\caption{The visualization results on ADE20K for 100-10 tasks after all CL steps. }
	\label{fig:vis-ade}
\end{figure}

Fine-grained incremental segmentation is one of the main challenges in continual segmentation tasks. Concretely, CPP achieves superior plasticity on new classes against the pioneer models on 100-50 task.  Compared to previous state-of-the-art method LAG, CPP+ also achieves effective improvement in both stability and plasticity. Quantitatively, on 100-50 task, CPP achieves the highest fine-grained performance on new classes. While on 100-10 task, CPP+ surpasses CPP on the initial classes by 1.28\% in mIoU and on the novel classes by 1.87\% in mIoU, respectively. On the other hand, the proposed model can alleviate semantic confusion among fine-grained classes. As seen in Fig.~\ref{fig:vis-ade}, CPP and CPP+ maintain integrate semantic boundary after all CL steps, validating the anti-forgetting ability.

\begin{table}[tbp]
	\caption{Quantitative evaluation on ADE20K valset in terms of mIoU (\%). The first and second best results are highlighted in bold and underlined, respectively.}
	\centering
	\setlength{\tabcolsep}{0.5mm}{
		{\begin{tabular}{l|c|ccc|ccc}
				\toprule[0.5mm]
				\multirow{2}*{Method}&\multirow{2}*{Model} &\multicolumn{3}{c}{100-50 (2 steps)}   &\multicolumn{3}{c}{100-10 (6 steps)} \\
				&& 1-100 & 101-150 & all & 1-100 & 101-150 & all \\
				\midrule
				MiB~\cite{MiB}&DeepLabv3&40.52&17.17&32.79&38.21&11.12&29.24 \\
				PLOP~\cite{PLOP}&DeepLabv3&41.87&14.89&32.94&40.48&13.61&31.59\\
				RCIL~\cite{RCIL}&DeepLabv3&42.30&18.80&34.50&39.30&17.60&32.10\\
				IDEC~\cite{IDEC}&DeepLabv3&42.01&18.22&34.08&40.25&17.62&32.71\\
				LAG~\cite{LAG}&DeepLabv3&41.64&19.73&34.34&41.00&18.69&33.56\\
				\cmidrule{1-8} 
				\textbf{CPP} &MaskFormer&\underline{42.86}&\textbf{25.32}&\underline{37.01} &\underline{41.99}&\underline{20.35}&\underline{34.78} \\
				\textbf{CPP+} &Mask2Former&\textbf{44.26}&\underline{24.82}&\textbf{37.78}&\textbf{43.27}&\textbf{22.22}&\textbf{36.25}\\
				\midrule
				\emph{offline} &Mask2Former&48.69&37.22&44.87&48.69&37.22&44.87 \\
				\bottomrule[0.5mm]
		\end{tabular}}}
	\label{table-ADE}
\end{table}

\begin{table}[tbp]
	\caption{Quantitative evaluation on COCO valset in terms of PQ (\%). $\dagger$ indicates we re-implement the model. }
	\centering
	\setlength{\tabcolsep}{1.5mm}{
		\begin{tabular}{c|l|ccc|ccc}
			\toprule[0.5mm]
			&\multirow{2}*{Method} &\multicolumn{3}{c}{40-40 (2 steps)}   &\multicolumn{3}{c}{40-10 (5 steps)} \\
			&&$C^{o}$&$C^{n}$&$C^{a}$&$C^{o}$&$C^{n}$&$C^{a}$ \\
			\midrule
			\multirow{3}*{\makecell{Seg\\only}}
			&\textbf{CPP} &\cellcolor{lightgray!50}32.26&\cellcolor{lightgray!50}26.59&\cellcolor{lightgray!50}29.48&\cellcolor{lightgray!50}27.26&\cellcolor{lightgray!50}17.35&\cellcolor{lightgray!50}22.31\\
			&\textbf{CPP+} &\cellcolor{cyan!30!purple!50}39.52&\cellcolor{cyan!30!purple!50}36.17&\cellcolor{cyan!30!purple!50}37.85&\cellcolor{cyan!30!purple!50}32.71&\cellcolor{cyan!30!purple!50}26.23&\cellcolor{cyan!30!purple!50}29.47\\
			&\emph{offline} &50.93&48.17&49.55&50.93&48.17&49.55\\
			\midrule
			\multirow{5}*{\makecell{Multi\\modal}}&\textbf{CPP} 
			& \cellcolor{lightgray!50}32.42&\cellcolor{lightgray!50}27.93&\cellcolor{lightgray!50}30.18&\cellcolor{lightgray!50}27.41&\cellcolor{lightgray!50}19.72&\cellcolor{lightgray!50}23.57\\
			&&\textcolor{darkgray}{$\uparrow$0.16}&\textcolor{darkgray}{$\uparrow$1.34}&\textcolor{darkgray}{$\uparrow$0.70}&\textcolor{darkgray}{$\uparrow$0.15}&\textcolor{darkgray}{$\uparrow$2.37}&\textcolor{darkgray}{$\uparrow$1.26}\\
			&\textbf{CPP+} &\cellcolor{cyan!30!purple!50}40.53&\cellcolor{cyan!30!purple!50}37.41&\cellcolor{cyan!30!purple!50}38.97&\cellcolor{cyan!30!purple!50}34.18&\cellcolor{cyan!30!purple!50}27.92&\cellcolor{cyan!30!purple!50}31.05\\
			&&\textcolor{purple!70}{$\uparrow$1.01}&\textcolor{purple!70}{$\uparrow$1.24}&\textcolor{purple!70}{$\uparrow$1.12} &\textcolor{purple!70}{$\uparrow$1.47} &\textcolor{purple!70}{$\uparrow$1.69} &\textcolor{purple!70}{$\uparrow$1.58}\\
			&\emph{offline} &51.79&48.93&50.36&51.79&48.93&50.36\\
			\bottomrule[0.5mm]
	\end{tabular}}
	\label{table-COCO}
\end{table}

\subsubsection{COCO} 
 We select samples from COCO those have pairwise segmentation and caption annotations. However, the caption annotations do not follow fine-grained class-wise format. Thus we evaluate the proposed model from two aspects: \textit{segmentation only} and  \textit{multi-modal}. The former only utilize segmentation annotations for CL training. While the latter utilize both segmentation and captioning annotation for CL training. As shown in Table~\ref{table-COCO}, we compare the segmentation under PQ metric. Considering the segmentation-only incremental tasks, the proposed CPP+ achieves superior performance on old, new and all average classes than that of CPP, which proves a stronger baseline model can boost multimodal CL performance.  With respect to cross-modal incremental learning tasks, CPP and CPP+  achieve significant improvements over single-modal methods by leveraging multimodal synergy. Specially for CPP+, the cross-modal fusion brings 1.58\% improvement on PQ in 40-10 task, proving the effectiveness of the multimodal collaboration.

\subsection{Ablation Study}
\label{sec:Abla}
\subsubsection{Module Contribution}
\label{sec:abla-MC}
To reveal the contribution of each module in the proposed method, we respectively disclose the corresponding modules in CPP and CPP+ as seen in Table~\ref{tab:abla-MC}. 
The proposed MCKD combines segmentation flow and captioning flow to perform KD across multimodal tasks.  To disclose the effectiveness of MCKD, the ICD and CID are separately and jointly verified, respectively. Compared to the fine-tuning approach, ICD achieves significant improvement, proving the anti-forgetting effectiveness of leveraging intermediate features constraints between $M^{t-1}$ and $M^t$. While CID emphasizes the consistency of output distribution at CL steps to align the ability from the old model. On the other hand, the synergy of ICD and CID proves the homogeneity in the proposed CPP for multimodal CL task. For example, in CPP task, the PQ on $C^a$ achieves 1.29\% higher than ICD only and 4.34\% higher than CID only on 15-5 task. For CPP+, the contribution of MCKD is also verified. On the other hand, the proposed CBC brings significant improvement to the segmentation performance by increasing 2.33\% (15-5) and 1.67\% (15-2) PQ on $C^a$, respectively.

Note that when not adopting SAPL, we apply a confidence-based pseudo-labeling with a fixed threshold with $\Gamma=0.7$. The detailed ablation study on pseudo-labeling is shown in Sec.~\ref{Sec-PL}.
\begin{table}[t]
	\centering
	\caption{Ablation study of segmentation performance of the proposed CPP and CPP+ in FineGrip evaluated by PQ (\%).}
	\label{tab:abla-MC}
	\setlength{\tabcolsep}{1.5mm}{
		\begin{tabular}{l|l|ccc|ccc}
			\toprule[0.5mm]
			&\multirow{2}*{Method} &\multicolumn{3}{c}{15-5 (3 steps)}&\multicolumn{3}{c}{15-2 (6 steps)}\\
			&&$C^{o}$ &$C^{n}$ &$C^{a}$ &$C^{o}$ &$C^{n}$ &$C^{a}$\\
			\midrule
			\multirow{5}*{CPP} &\textit{fine-tuning} &2.85&5.11&3.75 &1.28&0.77&1.08 \\
			&+ICD &27.88&23.07&25.96&16.22&9.37&13.48 \\
			&+CID &25.05&19.71&22.91&15.23&7.29&12.05 \\
			&+ICD\&CID &28.15&\textbf{25.89}&27.25&16.33&9.87&13.75 \\
			&+SAPL &\textbf{30.17}	&25.84 &\textbf{28.44}&\textbf{17.73}	&\textbf{10.62}&\textbf{14.89} \\
			\midrule
			\multirow{6}*{CPP+} &\textit{fine-tuning}&3.73&5.46&4.42&1.60&1.00&1.36\\
			&+ICD&30.29&27.13&29.03&23.21&17.79&21.10\\
			&+CID&28.53&24.95&27.10&22.52&19.24&21.21\\
			&+ICD\&CID&31.19&28.44&30.09&23.47&19.21&21.77\\
			&+CBC&\textbf{33.85}&30.28&32.42&24.82&\textbf{21.37}&23.44\\
		    &+SAPL&33.64&\textbf{31.51}&\textbf{32.79}&\textbf{25.70}&21.26&\textbf{23.92}\\
			\bottomrule[0.5mm]
		\end{tabular}}
\end{table}

\begin{table}[t]
	\centering
	\caption{Impact of hyper-parameters on FineGrip 15-5 task evaluated by PQ (\%).}
	\label{tab:Abla-weight}
	\begin{tabular}{c|ccccc}
		\toprule[0.5mm]
		$\lambda$&0.3&0.5&1.0&2.0&3.0\\
		\midrule
		CPP&27.12&27.65&27.77&\textbf{28.44}&27.69\\
		CPP+&31.79&32.71&\textbf{32.79}&32.53&32.05\\
		\bottomrule[0.5mm]
	\end{tabular}
\end{table}

\begin{figure}[t]
	\centering
	\includegraphics[scale=0.26]{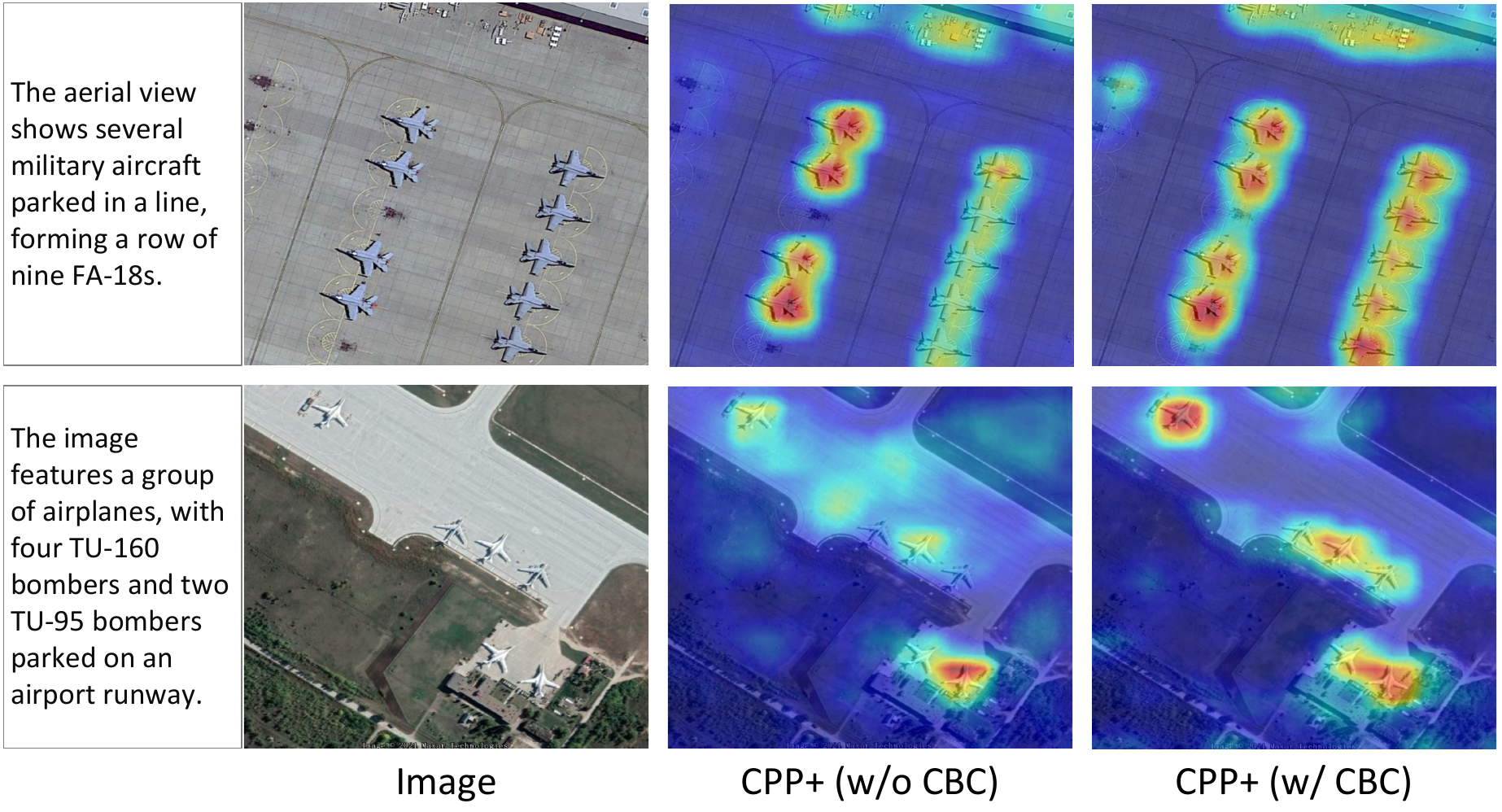}
	\caption{The attention map of the proposed CBC constraint. }
	\label{fig:vis-CBC}
\end{figure}

Specifically, the proposed CBC module enhances cross-modal interactive optimization and alignment. As seen in Fig.~\ref{fig:vis-CBC}, we investigate the cross-modal alignment efficiency during incremental steps. Prior to adopting CBC module, the feature distributions exhibit substantial dispersion with inadequate alignment between modalities. However, upon integration of the CBC module, the attention is more clustered on the foreground things, aligning with the fine-grained semantics in the text embedding. It demonstrates the enhanced alignment between corresponding image and text features in the embedding space. This observable improvement in feature coherence quantitatively validates the effectiveness of our proposed module in facilitating cross-modal alignment.

\begin{figure}[t]
	\centering
	\includegraphics[scale=0.31]{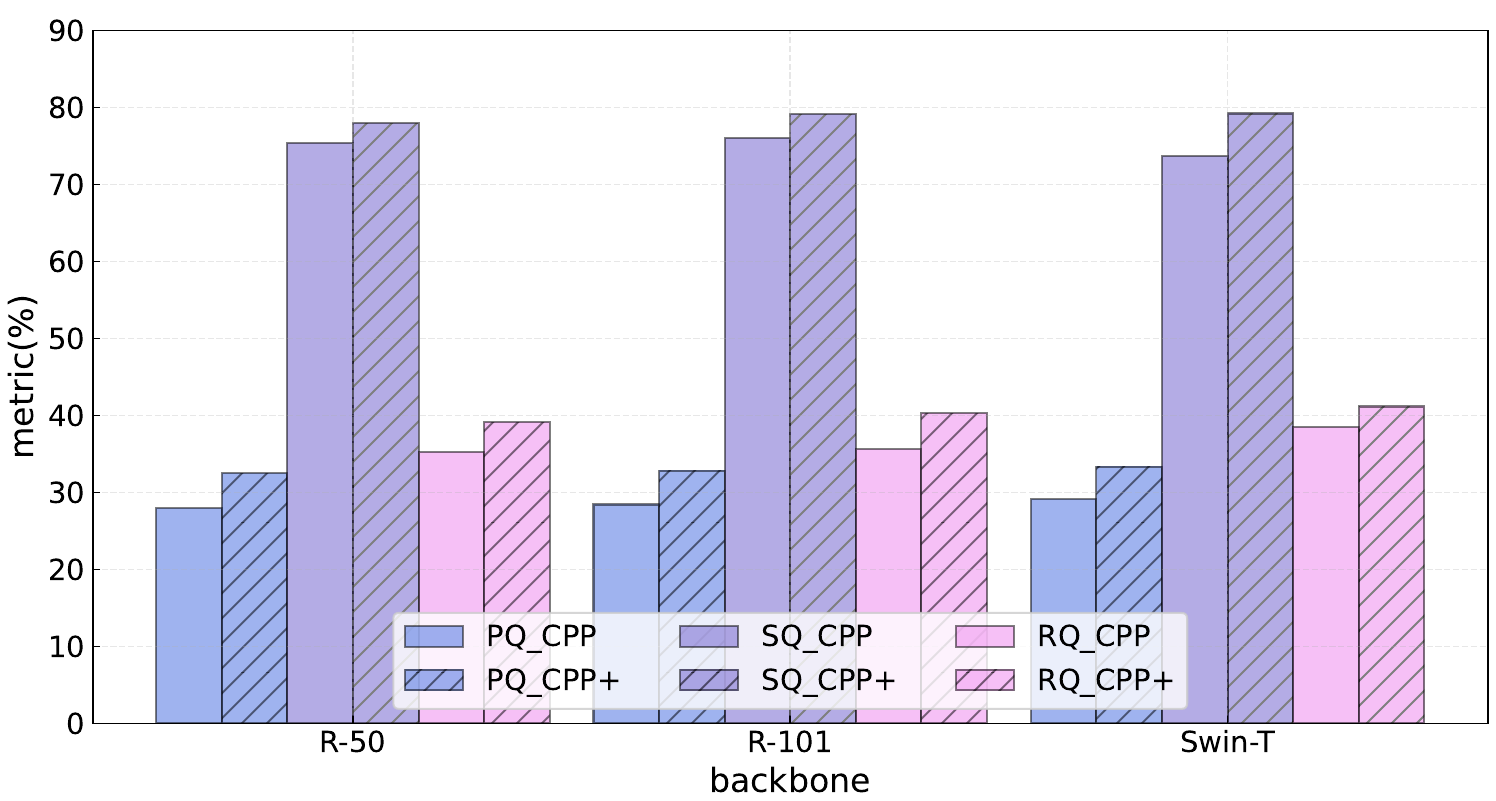}
	\caption{Comparison of PQ, SQ and RQ on all learned classes after all CL steps on FineGrip 15-5 task.}
	\label{fig:Abla-B}
\end{figure}
\begin{figure*}[t]
	\centering
 	\begin{subfigure}{0.45\textwidth}
 		\includegraphics[width=\linewidth]{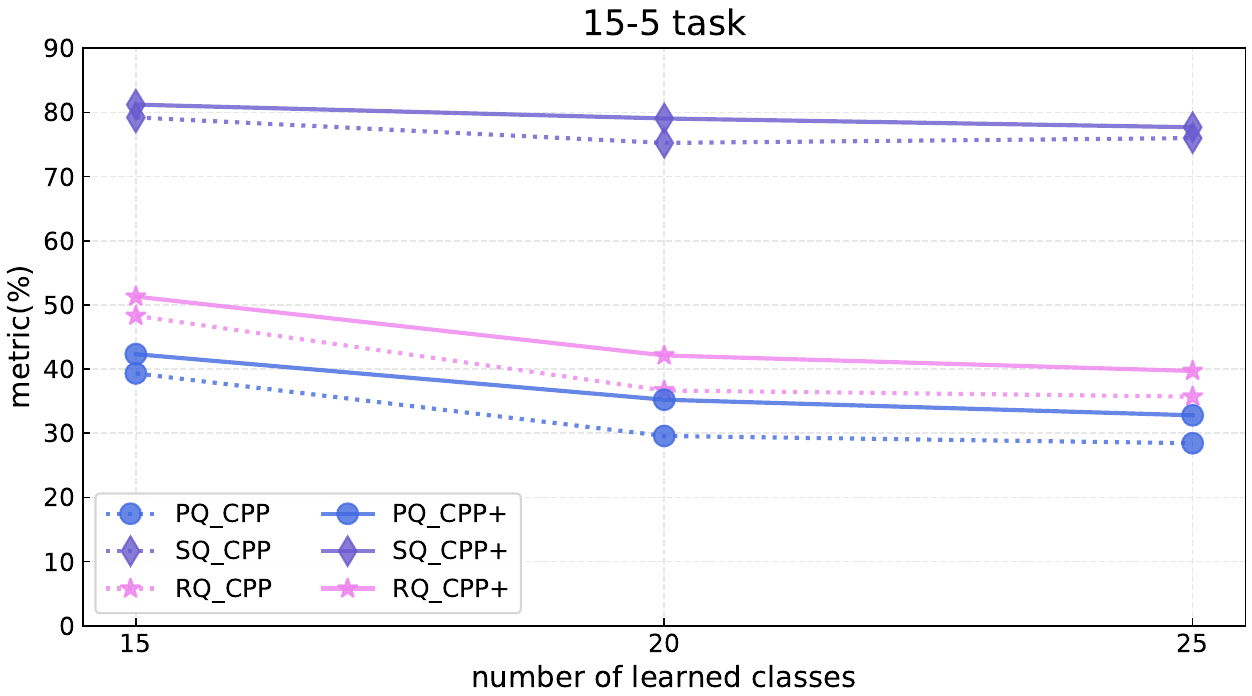}
 		\caption{}
 		\label{fig:Abla-AF-a}
 	\end{subfigure}
 	\hfill
 	\begin{subfigure}{0.45\textwidth}
 		\includegraphics[width=\linewidth]{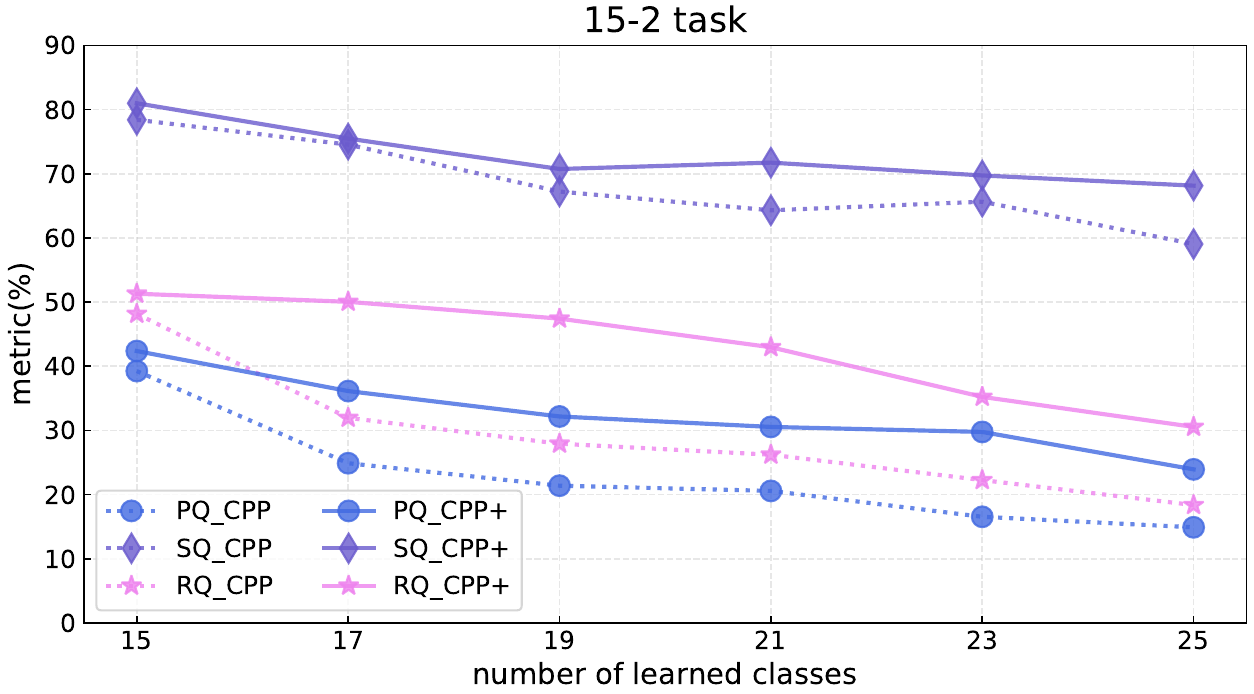}
 		\caption{}
 		\label{fig:Abla-AF-b}
 	\end{subfigure}
	\caption{The PQ, SQ and RQ evolution against the number of learned classes on FineGrip. }
	\label{fig:Abla-AF}
\end{figure*}

\subsubsection{Impact of hyper-parameters}
As defined in Eqn.~\ref{eqn-overall},  CPP utilizes a multi-task learning approach within a unified architecture. The weights assigned to multimodal tasks significantly affect the efficiency of CL. As shown in Table~\ref{tab:Abla-weight}, the performance of CPP achieves the highest PQ since the captioning task is weighted as $\lambda=2.0$, demonstrating the cross-modal boosting in CPP tasks. While CPP+ achieves the highest performance under $\lambda=1.0$. We believe this is because CPP+ is designed to place more emphasis on segmentation tasks, which are critical for achieving fine-grained scene understanding. In summary, the analysis of $\lambda$ reveals that task weighting plays a crucial role in balancing multimodal interactions and achieving optimal performance for multimodal CL tasks.

\subsubsection{Impact of base model}
The proposed architecture is model-agnostic, which supports various backbones and networks. For the proposed CPP and CPP+, we propose a hypothesis that a strong base model can improve the CL efficiency. As seen in Fig.~\ref{fig:Abla-B}, various backbones including ResNet-50~\cite{resnet}, ResNet-101 and Swin-T~\cite{swin-T} are explored. The results can be analyzed from two aspects. On the one hand, CPP+ utilizes a stronger baseline than that of CPP, proving the effectiveness by a large margin in PQ.
Quantitatively, the model with ResNet-101 achieves higher PQ, SQ and RQ than that with ResNet-50, which proves a stronger backbone can achieve higher anti-forgetting performance and compatibility on new classes. It also shows that Transformer-based models achieve better recognition performance since the RQ score is significantly higher than two CNN-based models. However, CNN-based models achieve higher segmentation performance as the SQ scores are higher. We prefer it since the local features are more attentively utilized especially for small objects and areas.

\subsubsection{Compatibility on plasticity and stability} 
With the incremental arriving data, the performance of old classes degraded since lacking old annotations. To validate the anti-forgetting ability of the proposed CPP model, we evaluate the PQ, SQ and RQ evolution against the number of learning classes on FineGrip 15-5 task and 15-2 task, respectively. As seen in Fig.~\ref{fig:Abla-AF}, For CPP, the PQ and RQ performance of the old classes suffer severe degradation primarily attributed to catastrophic forgetting and semantic drift.  Fig.~\ref{fig:Abla-AF}(a) shows the model degeneration after all incremental training steps, which is the reflection of the stability-plasticity dilemma game. For FineGrip 15-2 task, we observe two interesting but valuable aspects from Fig.~\ref{fig:Abla-AF}(b). On the one hand, SQ is higher than that of RQ. This indicates a fundamental limitation in preserving fine-grained image-level captioning under incremental learning scenarios. In contrast, SQ maintains a high score since the background classes occupy the most pixels and being learned at the initial step.This observation highlights that recognition tasks in multimodal CL task inherently resist forgetting due to their reliance on coarse-grained semantic consistency. On the other hand, the SQ evolution appears to be achieve higher performance when unknown semantics is learned step by step.

While the proposed CPP+ architecture demonstrates markedly improved robustness. The model maintains higher PQ, SQ and RQ scores, suggesting enhanced preservation of cross-modal synergistic interaction. Here we propose an intriguing conclusion, that is fine-grained semantic recognition (RQ) remains vulnerable to incremental shifts, instance recognition (SQ) benefits significantly from semantic stability. This aligns with our hypothesis that recognition tasks rely more on global and fine-grained feature relationships rather than single-task pixel dependencies.

\subsubsection{Impact of pseudo-labeling} 
\label{Sec-PL}
Pseudo-labeling serves as a promising strategy to alleviate catastrophic forgetting in exemplar-free conditions. We compare the proposed SAPL with the typical single-task confidence-based method with a \textit{fixed} threshold, in which we set $\Gamma=0.5$ as a convention and $\Gamma=0.7$ following~\cite{IDEC}. As seen in Table~\ref{tab:Abla-PL}, for CPP, the proposed SAPL achieves 2.02\% PQ improvement on $C^{o}$ and 1.19\% superiority on $C^a$ to the fixed confidence threshold setting ($\Gamma=0.7$), which validates that cross-task interaction mechanisms enhance pseudo-label reliability by dynamically balancing historical and incremental knowledge. However, the decline in $C^n$ shows the conflict between retaining the old knowledge and bias on the new knowledge. While for CPP+, the proposed SAPL boosts $C^n$ by 1.23\% and $C^a$ by 0.37\% PQ, respectively. Nevertheless, the observed 0.21\% PQ decline on $C^o$ and 0.12\% degradation on $B^{a}$ than that of $\Gamma=0.7$ indicate the intricate trade-off in multimodal CL tasks,  where modality-specific feature drifts and task heterogeneity amplify optimization conflicts..

\begin{table}[t]
	\centering
	\caption{Comparison of various pseudo-labeling methods on FineGrip 15-5 task.}
	\label{tab:Abla-PL}
	\setlength{\tabcolsep}{1.5mm}{
	\begin{tabular}{c|c|ccc|cc}
		\toprule[0.5mm]
		&\multirow{2}*{Method} &\multicolumn{3}{c}{PQ}&\multicolumn{2}{c}{BLEU}\\
		&&$C^{o}$ &$C^{n}$ &$C^{a}$ &$B^{b}$&$B^{a}$ \\
		\midrule
		\multirow{3}*{CPP}&fixed ($\Gamma=0.5$)&27.93&25.30&26.88 &33.01&25.38\\
		&fixed ($\Gamma=0.7$)&28.15&\textbf{25.89}&27.25 &33.01&\textbf{27.40}\\
		&SAPL&\textbf{30.17}&25.84&\textbf{28.44} &33.01&27.00\\
		\midrule
		\multirow{3}*{CPP+}&fixed ($\Gamma=0.5$)&33.15&30.42&32.06&38.63&\textbf{36.49} \\
		&fixed ($\Gamma=0.7$)&\textbf{33.85}&30.28&32.42&38.63&36.44 \\
		&SAPL&33.64&\textbf{31.51}&\textbf{32.79}&38.63&36.32 \\
		\bottomrule[0.5mm]
	\end{tabular}}
\end{table}

\begin{table}[t]
	\centering
	\caption{Average performance on various class incremental orders on FineGrip 15-5 task in terms of PQ (\%).}
	\label{table:ClassOrders}
	\setlength{\tabcolsep}{1.0mm}{
		\begin{tabular}{c|c|cc|c} 
			\toprule[0.5mm]
			Method&Order&$C^{1:10\&21-25}$&$C^{11:20}$&$C^{1:25}$\\
			\midrule
			\multirow{6}*{CPP}&a&30.17&25.84&28.44\\
			&b&25.79&24.15&25.13\\
			&c&28.73&24.96&27.22\\
			&d&22.57&27.18&24.41\\
			&e&28.32&22.29&25.91\\
			&avg.$\pm$std.&27.12$\pm$2.68&24.88$\pm$1.64&26.22$\pm$1.45\\
			\midrule
			\multirow{6}*{CPP+}&a&33.64& 31.51&32.79 \\
			&b&31.28 &29.50 &30.57\\
			&c&32.17 &30.29 &31.42\\
			&d&27.82 &32.17 &29.56\\
			&e&29.15 &33.21 &30.77\\
			&avg.$\pm$std.&30.81$\pm$2.09 &31.34$\pm$1.32&31.02$\pm$1.07\\
			\bottomrule[0.5mm]
	\end{tabular}}
\end{table}

\begin{figure}[ht]
	\centering
	\includegraphics[scale=0.36]{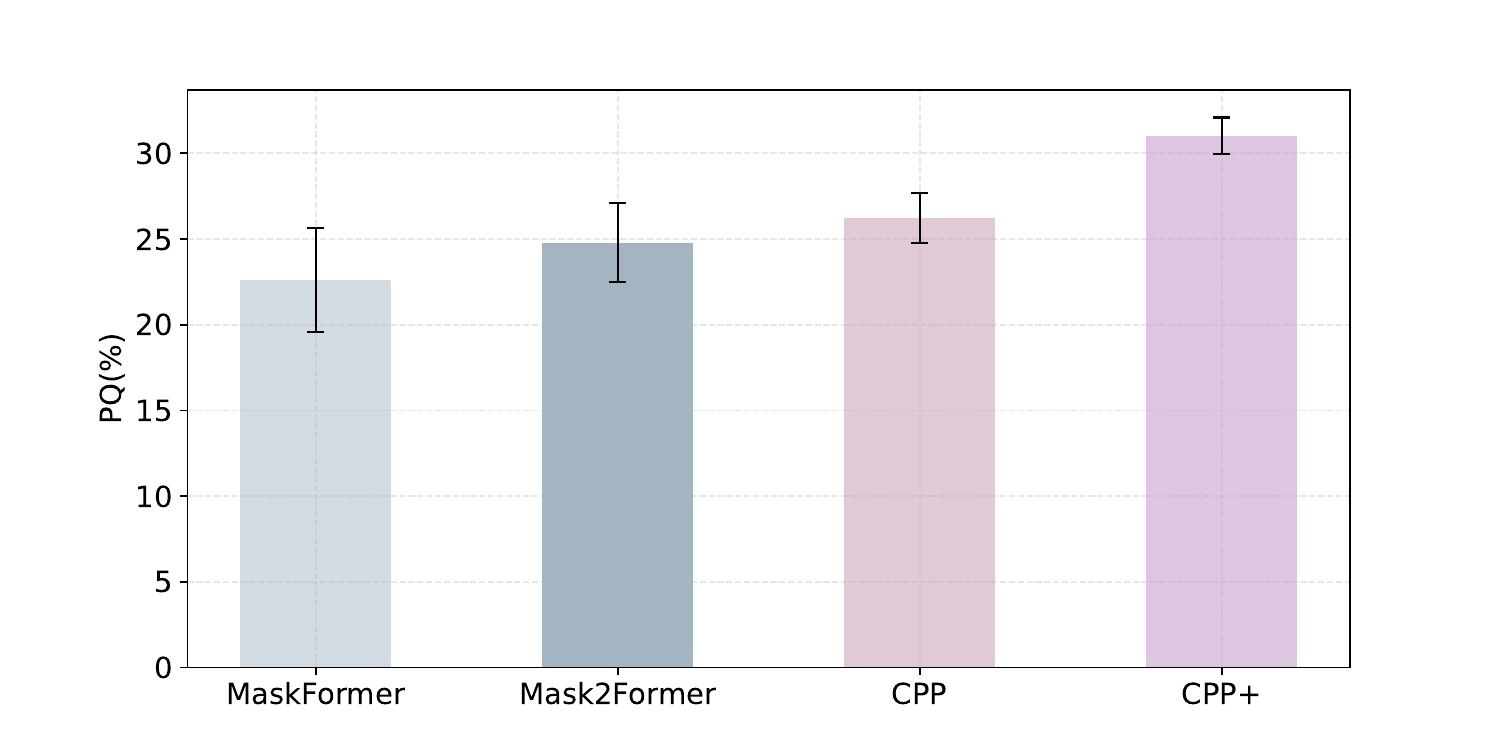}
	\caption{The average performance and standard variance under different incremental class orders on FineGrip 15-5 task.}
	\label{fig:avg+std}
\end{figure}

\begin{table}[t]
	\centering
	\caption{Computational complexity in CPP task.}
	\begin{tabular}{c|ccc}
		\toprule[0.5mm]
		Task&Params. (M)&FLOPs (G)&FPS\\
		\midrule
		Seg.-only&45.0&126&8.5\\
		Cap.-only&31.2&87&8.0 \\
		CPP&52.1& 135 &4.8\\
		CPP+&62.2&149&4.2\\
		\bottomrule[0.5mm]
	\end{tabular}
	\label{tab:ce}
\end{table}

\subsubsection{Robustness analysis}
To reveal the robustness to class learning orders of the proposed method, we perform experiments on FineGrip 15-5 task with five different class orders including the ascending order and four random orders on \textit{things} classes as follows. Specially, in FineGrip~\cite{FineGrip} dataset, $C^{21-25}$ indicates the background stuff classes, which serves as the partial base classes that are learned at the initial step in the CPP experiments. 
\begin{equation}
	\scriptsize
	\begin{split}
		\nonumber
		a:\{[21-25,1,2,3,4,5,6,7,8,9,10],[11,12,13,14,15],[16,17,18,19,20]\} \\
		b:\{[21-25,5,7,8,9,12,14,15,16,19,20],[1,2,4,11,13],[3,6,10,17,18]\} \\ 
		c:\{[21-25,3,4,5,8,9,13,15,17,19,20],[7,10,11,16,18],[1,2,6,12,14]\} \\ 
		d:\{[21-25,1,2,3,11,12,14,15,16,18,20],[7,8,10,17,19],[4,5,6,9,13]\} \\
		e:\{[21-25,2,3,5,7,9,10,12,13,14,19],[1,4,8,16,17],[6,11,15,18,20]\} \\
	\end{split}
\end{equation}

The results shown in Table~\ref{table:ClassOrders} indicate the PQ of $C^{1:10\&21-25}$ and $C^{11:20}$ after all CL steps. The average PQ and standard variance are reported. It reveals that the class incremental orders have an evident impact on CPP performance. For example, the average PQ on $C^{1:10\&21:25}$ varies sharply. It proves the critical challenge of catastrophic forgetting in multi-task CL. On the other hand, the performance in all classes after all CL steps disclose the learning stability. As shown in Fig.~\ref{fig:avg+std}, CPP and CPP+ achieves  more stable incremental learning compared to the base model in multimodal incremental learning scenarios, proving the effectiveness of the proposed method.  

\subsubsection{Computational complexity}
Since the multimodal branches are conducted to perform continual panoptic perception, we reveal the computational complexity of the proposed model compared with single-task CL approaches. The results in Table~\ref{tab:ce} come from 20-5 task with 800$\times$800$\times$3 input size after all CL steps. The result indicates CPP and CPP+ implements multi-task and multimodal CL with extra cost than single-task CL approaches, which is accompanied by a reducing of inference efficiency.  Despite this trade-off, the benefits of CPP in terms of comprehensive perception and adaptability to evolving tasks justify the additional computational cost. The ability to perform multi-task and multimodal learning in a unified framework enables CPP to achieve superior performance in complex, real-world scenarios where single-task approaches may fall short. However, optimizing the computational efficiency of CPP remains an important area for future work.

\subsubsection{Limitations}
\begin{figure}[tbp]
	\centering
	\includegraphics[scale=0.5]{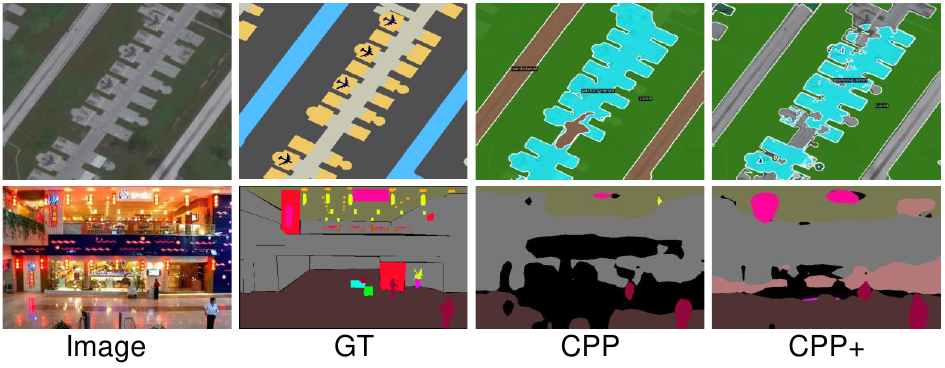}
	\caption{Failure examples of the proposed model.}
	\label{fig:failure}
\end{figure}

In Fig.~\ref{fig:failure}, we display several failure examples of the proposed method. The problem is centered in complex scenes and inter-class confusion among extremely abundant semantics coexistence.  On the one hand, the shackle of multimodal incremental learning lays in semantic chaos. On the other hand, the potential improvement of CPP is to establish the strict coupling and mutual verification between the multimodal information during incremental training steps. In addition, considering CPP from multi-task learning perspective, the seesaw phenomenon is a non-negligible issue since pixel-level segmentation and image-level captioning tasks have different training difficulty and convergence efficiency. 

\section{Conclusion}
In this paper, we propose continual panoptic perception (CPP), a multimodal continual learning approach for a comprehensive perception for continually accumulated image data and evolving tasks. The core contribution lies in unifying cross-modal alignment, knowledge retention, and semantic consistency into a unified continual learning paradigm. This integration enables the model to maintain robust performance across a wide range of tasks while adapting to new data and modalities over time.  Extensive experiments on three challenging benchmarks demonstrate CPP’s superiority over state-of-the-art methods, highlighting its effectiveness in handling complex, multimodal continual learning scenarios. 

Despite these advances, the cross-modal task harmonization and generalization in open-world contexts are still potential challenges for CPP task. And the investigation of universal CL manner for extremely abundant modalities is also a promising but challenging prospect. Our future work will focus on advancing unified intelligent systems that resolving cross-modal forgetting patterns through universal and applicable architectures.

\ifCLASSOPTIONcaptionsoff
  \newpage
\fi

{\small
	\bibliographystyle{IEEEtran}
	\bibliography{1}
}

\end{document}